\documentclass{article}

\PassOptionsToPackage{numbers, square}{natbib}
\usepackage[final]{nips_2016} 
\usepackage[utf8]{inputenc} % allow utf-8 input
\usepackage[T1]{fontenc}    % use 8-bit T1 fonts
\usepackage{hyperref}       % hyperlinks
\usepackage{url}            % simple URL typesetting
\usepackage{booktabs}       % professional-quality tables
\usepackage{amsfonts}       % blackboard math symbols
\usepackage{microtype}      % microtypography
\usepackage{graphicx}		% for /includegraphics
\usepackage{caption}		% for stacking images
\usepackage{subcaption}		% for stacking images
\usepackage{xcolor}
\usepackage{amsmath}
\DeclareMathOperator*{\argmax}{argmax}
\DeclareMathOperator*{\argmin}{argmin}
\usepackage{lmodern}
\usepackage[T1]{fontenc}
\usepackage{textcomp}
\usepackage{tcolorbox}
\usepackage{fancybox}
\usepackage{varwidth}
\usepackage{algorithm}
\usepackage[noend]{algpseudocode}
\tcbuselibrary{raster}
\tcbuselibrary{breakable}
\newcommand\tab[1][1.70cm]{\hspace*{#1}}

\makeatletter
\def\BState{\State\hskip-\ALG@thistlm}
\makeatother

\usepackage[normalem]{ulem}	% for strike-out text use \sout{text}
\usepackage[toc,page]{appendix}	%for appendices
\begin{document}

% TODO: April 9th: We'll change the title and use something catchy and 
\title{ReGAN: \underline{RE}[LAX|BAR|INFORCE] based Sequence Generation using \underline{GAN}s}

% The \author macro works with any number of authors. There are two
% commands used to separate the names and addresses of multiple
% authors: \And and \AND.
%
% Using \And between authors leaves it to LaTeX to determine where to
% break the lines. Using \AND forces a line break at that point. So,
% if LaTeX puts 3 of 4 authors names on the first line, and the last
% on the second line, try using \AND instead of \And before the third
% author name.

% \author{
%   Aparna Balagopalan, Mathieu Ravaut, Raeid Saqur, Satya Gorti \thanks{Equal Contribution by the Authors} \\
%   \\
%   Department of Computer Science\\
%   University of Toronto\\
%   \texttt{[aparna, mravox, raeidsaqur, satyag]@cs.toronto.edu} 
% }
\author{
  Aparna Balagopalan \thanks{All authors contributed equally}\\
  Department of Computer Science\\
  University of Toronto\\
  \texttt{aparna@cs.toronto.edu} \\
  \And
  Satya Krishna Gorti \footnotemark[1] \\
  Department of Computer Science \\
  University of Toronto \\
  \texttt{satyag@cs.toronto.edu} \\
  \And
  Mathieu Ravaut \footnotemark[1]\\
  Department of Computer Science \\
  University of Toronto \\
  \texttt{mravox@cs.toronto.edu} \\
  \And
  Raeid Saqur \footnotemark[1]\\
  Department of Computer Science \\
  University of Toronto \\
  \texttt{raeidsaqur@cs.toronto.edu} \\
}
%\thanks{All authors contributed equally}

%test

\maketitle
\begin{abstract}
%TODO: RS: Rewrite this, highlight the main contibutions.
Generative Adversarial Networks (GANs) have seen steep ascension to the peak of ML research zeitgeist in recent years. Mostly catalyzed by its success in the domain of image generation, the technique has seen wide range of adoption in a variety of other problem domains. Although GANs have had a lot of success in producing more realistic images than other approaches, they have only seen limited use for text sequences. Generation of longer sequences compounds this problem. Most recently, SeqGAN (Yu et al., 2017) has shown improvements in adversarial evaluation and results with human evaluation compared to a MLE based trained baseline. The main contributions of this paper are three-fold: 1. We show results for sequence generation using a GAN architecture with efficient policy gradient estimators, 2. We attain improved training stability, and 3. We perform a comparative study of recent unbiased low variance gradient estimation techniques such as REBAR (Tucker et al., 2017), RELAX (Grathwohl et al., 2018) and REINFORCE (Williams, 1992). Using a simple grammar on synthetic datasets with varying length, we indicate the quality of sequences generated by the model.
\end{abstract}

\section{Introduction}\label{introduction}
 The task of meaningful text generation is of crucial interest. To generate long sequences that make sense from end to end, a machine learning model must learn to keep and manage a lot of context and abstract features. In the past years, RNN-based neural networks have been trained successfully to predict the next character or word from a sequence of previous characters or words. However, directly producing full sequences, in natural language for instance, remains an active research area. 

Recently, Generative Adversarial Networks (GANs) \cite{goodfellow2014generative} have shown great performance on several generation tasks, especially image generation \cite{denton2015deep}. However, less work has been done regarding the task of generating sequential data, mainly because of two structural aspects of GANs. First, in GANs, the generator gets updated by backpropagating gradients from the outputs of the discriminator, which makes little sense when the generated data is made of discrete tokens. Secondly, GANs can only give a score once an entire sequence has been generated, which is a problem with large sequences of tokens.

To alleviate these two issues, SeqGAN \cite{yu2017seqgan} takes a reinforcement learning approach while generating tokens, and considers the generative model as a stochastic parametrized policy. This way, gradients can pass back from the discriminator loss to the generator’s weights. When a token is generated, the rest of the sequence is drawn with Monte Carlo sampling, so that the model bypasses the need to wait for a full sequence generation to get a score. It can get a score at each token generation via averaging these Monte Carlo samples. The reward is given by the discriminator's output when fed with the generated sequence. In this paper, we follow a similar approach and use the discriminator for estimating the reward function.

\section{Related Work}\label{related}
% SeqGAN
We briefly discuss a few developments in generative modeling of text, especially the works that utilize policy gradient estimators in their approach.

% \subsection{Models for Text Generation} % Only one subsection removing.
Early papers for text generation such as \cite{guo2015generating} use sequence to sequence modeling with Deep Q-learning to generate natural language sentences. This is done by constructing an LSTM encoder-decoder network to first encode an input sentence and a Deep Q-Network to take decoding decisions pertaining to the output sequence in an iterative manner. An attention-based 'difficulty' measure is implemented to force the DQN to explore the whole discrete sequence space (for example difficult words from a list of words to choose from).

In GAN-based settings such as SeqGAN \cite{yu2017seqgan}, the discriminator outputs a 'reward' for a sequence of tokens/words generated by the generator. The reward in SeqGAN \cite{yu2017seqgan} is maximized with respect to the parameters of the generator using policy gradients. In contrast to our work, the discriminator in this case is able to see the token transition probabilities involved in the generation of a sequence as the reward function incorporates these probabilities.

In MaskGAN \cite{fedus2018maskgan}, reward is generated at every token of the sequence by training the text generation model to fill in missing text conditioned on the surrounding context, yielding more informative signals to the generator. An actor-critic training procedure is followed as the critic can help reduce the high variance of the gradient updates (computed using REINFORCE \cite{williams1992simple}) in action space. They specify a cost function on the output of the generator that encourages high sample quality.

Other discrete GANs such as MaliGAN \cite{che2017maximum}, in contrast to our work, use importance sampling to build a novel objective for the generator to optimize. The intuition behind this is that gradients with lower variance are obtained with an objective function closer to maximum likelihood
(MLE) training of auto-regressive model. A novel objective has been proposed in \cite{zhang2017adversarial} as well, focusing on alleviating the mode collapse problem. This is done by using a kerneled discrepancy matrix for matching the high-dimensional latent feature distributions of real and synthetic sentences.

The discrete space issue is side-stepped without using gradient estimators in \cite{rajeswar2017adversarial} by letting the discriminator see a sequence of probabilities over every token in the vocabulary from the generator and a sequence of one-hot vectors from the true data distribution.

For a more detailed discussion of related work pertaining to gradient estimators, please refer to apppendix \ref{appendix:background-gradient-estimators}.

\newpage

\section{Sequential GAN using gradient estimators}\label{algo}
%TODO: Add our training algorithm
Mapping from the previous section to the context of a GAN, notations now have the following meaning:
\begin{itemize}
\item \(b\) is the sequence being sampled.
\item $\theta$ parametrizes the generator. 
\item Our reward is given by the discriminator's output. Thus $f(b)$ is the output of the discriminator \emph{on a whole sequence}.
\end{itemize}

As usual, we train the GAN via mini-batch training. We introduce the following training parameters:
\begin{itemize}
\item \(T\) is the sequence length.
\item $S_{1:T} = S_{T}$ is a full generated sequence, consisting of $s_{1}...s_{i}...s_{T}$. For $1 \leq t \leq T$, we note $S_{t}$ the partial sequence $S_{1:t}=s_{1}..s{t}$.
\item \(B\) is the batch size.
%  \item The reward is given by $R = D(G(z))$ where z is noise distribution given as input to the generator
%  \item We note $R = D(G(x))$ where x is some input noise the discriminator's output, as it corresponds to the reward function. We note $R_{j}$ the reward associated with an element j of the mini-batch.
\end{itemize}

We formulate the output of the discriminator as the reward which indicates the quality of the sentence generated by the generator. Hence we can denote the reward as $R = D(G_{\theta}(\zeta))$ where $\zeta$ is sampled from a noise distribution given as input to the generator.

Now our objective is to train the parameters of the generator to maximize the expected reward. The objective function is represented as:
\begin{equation}
J(\theta) = \mathbb{E}[D(S_{1:T})]
\end{equation}
Which can be rewritten as:
\begin{equation}
J(\theta) = \mathbb{E}[R|\theta]
\end{equation}

We need to optimize the parameters of the generator to maximize this expected reward, while we need to optimize the parameters of the discriminator to distinguish between real and synthetic data.

Our optimization problem to update the generator now is equivalent to solving:
% Following SeqGAN, the objective of the generator is to generate a sequence from end to end that maximizes the probability of fooling the discriminator. We note the generator loss function: 
% \begin{equation}
% J(\theta) = \mathbb{E}[R|\theta]
% \end{equation}
% And our optimization problem is equivalent to solving:
\begin{equation}
\hat{\theta} = \argmax_\theta J(\theta) = \argmax_\theta \mathbb{E}[R|\theta] = - \argmin_\theta \mathbb{E}[R|\theta]
\end{equation}

Since the input to the discriminator is obtained by sampling from softmax	 distribution produced by the generator, we cannot differentiate the output of the discriminator with respect to parameters of generator $\theta$. We hence need gradient approximation to backpropagate gradients.

\subsection{REINFORCE}
We first introduce REINFORCE to obtain an approximated, unbiased gradient of $J(\theta)$. Contrasting our method with SeqGAN \cite{yu2017seqgan}, SeqGAN gives a reward per generated token via Monte Carlo sampling to estimate the quality of choosing this token. We decided not to follow this approach. Instead, we write the REINFORCE estimator as follows:

\begin{align}
\nabla_{\theta} J_{RF}(\theta) &= \mathbb{E}_{S_{1:T} \sim G_{\theta}}[R\nabla_{\theta}\log(P_{\theta}(S_{T}))]\\
&= \sum_{j=1}^{B} R_{j} \nabla_{\theta}\log(P_{\theta}(S_{1:T})) \text{ when decomposing over the batch}
\end{align}
where $P_{\theta}$ is the probability distribution over tokens at the output of the generator. \\
We can decompose the sequence probability as follows:
\begin{equation}
P_{\theta}(S_{1:T}) = P_{\theta}(S_{1})\prod_{i=2}^{T}P_{\theta}(S_{i}|S_{i-1})
\end{equation}
Substituting the log of our previous expression, the REINFORCE gradient estimator then becomes:
\begin{equation}
\nabla_{\theta} J_{RF}(\theta) = \sum_{j=1}^{B} R_{j} \nabla_{\theta}[log(P_{\theta}(S_{1})) + \sum_{i=2}^{T}log(P_{\theta}(S_{i}|S_{i-1}))]
\end{equation}
\begin{figure}[!htbp]
\begin{center}
\includegraphics[width=10cm]{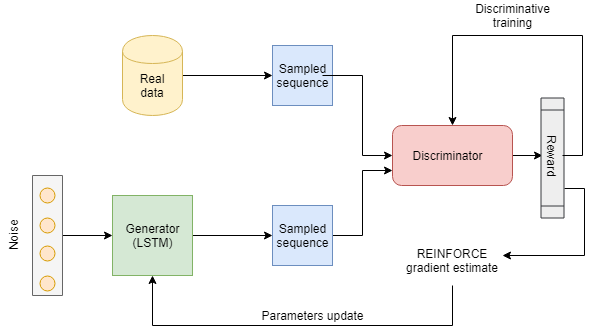}
\end{center}
\caption{Topological Architecture for REINFORCE}\label{reinforce-architecture}
\end{figure}

\subsection{REBAR}
To get the REBAR gradient estimator, following from \cite{grathwohl2017backpropagation}, one needs to compute $z \sim p(z|\theta)$, as well as a relaxed input conditioned on the discrete variable $s_{i}$, $\tilde{z} \sim p(z|s_{i},\theta)$. We also introduce a temperature parameter $\lambda$ that parameterizes a softmax $\sigma$ of these two variables $z$ and $\tilde{z}$, before computing $D(\sigma_{\lambda}(z))$ and $D(\sigma_{\lambda}(\tilde{z}))$.\\

\begin{figure}[!htbp]
\begin{center}
\includegraphics[width=10cm]{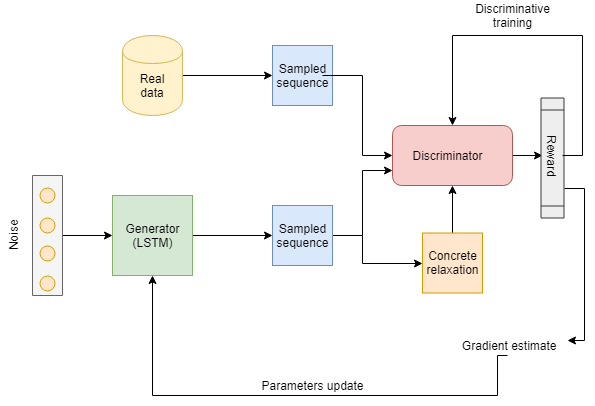}
\end{center}
\caption{Topological Architecture for REBAR}\label{rebar-architecture}
\end{figure}

The REBAR gradient estimator is written as follows:
\begin{align}
\nabla_{\theta} J_{RB}(\theta) = &\sum_{j=1}^{B} (R_{j} - D(\sigma_{\lambda}(\tilde{z_{j}}))) \nabla_{\theta}[log(P_{\theta}(S_{1})) + \sum_{i=2}^{T}log(P_{\theta}(S_{i}|S_{i-1}))] \\
+ &\frac{1} {B}\sum_{j=1}^{B}\nabla_{\theta}D(\sigma_{\lambda}(z_{j})) - \frac{1}{B}\sum_{j=1}^{B}\nabla_{\theta}D(\sigma_{\lambda}(\tilde{z_{j}}))
\end{align}
where the subscript \(j\) is relative to the current element of the batch. 

% \begin{figure}[!htbp]
%   \includegraphics[width=130pt, height = 110 pt]{REINFORCE_final.png}
%   \includegraphics[width=130pt, height = 110 pt]{REBAR_final.png}
%   \includegraphics[width=130pt, height = 110 pt]{RELAX.png}
% \caption{Sequential GAN training with REINFORCE, REBAR, RELAX (from left to right)}
% \end{figure}

\subsection{RELAX}
The last gradient estimator that we consider is RELAX. For this purpose, we introduce a third neural network which is a convolutional neural network  in our training, that we parameterize by $\phi$ and note $\hat{c_{\phi}}$. Following the notation of \cite{grathwohl2017backpropagation}, we note:
\begin{equation}
c_{\lambda, \phi}(z) = D(\sigma_{\lambda}(z)) + \hat{c_{\phi}}(z)
\end{equation}

\begin{figure}[!htbp]
\begin{center}
\includegraphics[width=10cm]{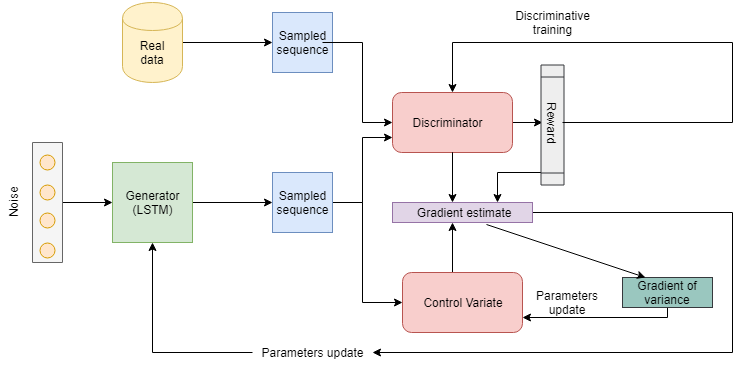}
\end{center}
\caption{Topological Architecture for RELAX}\label{relax-architecture}
\end{figure}

The RELAX gradient estimator of $J(\theta)$ is expressed:
\begin{align}
\nabla_{\theta} J_{RX}(\theta) = &\sum_{j=1}^{B} (R_{j} - c_{\lambda, \phi}(\tilde{z_{j}})) \nabla_{\theta}[log(P_{\theta}(S_{1})) + \sum_{i=2}^{T}log(P_{\theta}(S_{i}|S_{i-1}))] \\
&+ \frac{1} {B}\sum_{j=1}^{B}\nabla_{\theta}c_{\lambda, \phi}(z_{j}) - \frac{1}{B}\sum_{j=1}^{B}\nabla_{\theta}c_{\lambda, \phi}(\tilde{z_{j}})
\end{align}

In terms of implementation, the difference between the REBAR and RELAX gradient estimations simply consists of adding a term to the discriminator's output of the relaxation . \\

Finally, from an implementation point of view, RELAX requires getting gradients from each element of the batch. We are also backpropagation through time via the loop over \(i\) in all three gradient estimators, so training involves a double for loop, causing adversarial training to be slower than MLE pre-training of the generator.

Algorithm \ref{alg:the_alg} gives the flow of training a GAN using RELAX gradient estimator \footnote[1]{All code related to the project can be found at: \textit{https://github.com/TalkToTheGAN/REGAN}}.
\begin{algorithm}
\caption{Sequential GAN trained with RELAX}\label{alg:the_alg}
\begin{algorithmic}[1]
\State \textbf{arguments}: Generator $G_{\theta}$, Discriminator $D_{\phi}$, annex neural network $\hat{c_{\phi}}$
\State \textbf{initialize} all three networks randomly
\State \textbf{pre-train} the generator for \emph{PRE\_EPOCH\_GEN} epochs using MLE
\State \textbf{repeat}
\State \hspace{10mm} \textbf{for} g-steps \textbf{do}
\State \hspace{20mm} \textbf{Generate} \emph{BATCH SIZE} sequences from $G_{\theta}$ 
\State \hspace{20mm} Get their \textbf{rewards} via the discriminator
\State \hspace{20mm} Get the token probability distribution $p(b|\theta) \sim G_{\theta}$
\State \hspace{20mm} Compute \textbf{relaxations} $z$ and $\tilde{z}$ as in \cite{grathwohl2017backpropagation}
\State \hspace{20mm} Compute control variates $c_{\phi}(z)$ and $c_{\phi}(\tilde{z})$ as in (10)
\State \hspace{20mm} Get a gradient approximation for the generator $J_{RX}(\theta)$ as in (11) and (12)
\State \hspace{20mm} Get expected values of gradients over mini-batch
\State \hspace{20mm} \textbf{Update} $G_{\theta}$
\State \hspace{20mm} Get the \textbf{variance} of the gradient as in \cite{grathwohl2017backpropagation}
\State \hspace{20mm} Train $\hat{c_{\phi}}$ to minimize this variance
\State \hspace{10mm} \textbf{for} d-steps \textbf{do}  
\State \hspace{20mm} Use current $G_{\theta}$ to generate negative examples
\State \hspace{20mm} Train $D_{\phi}$ to discriminate between fake and real sequences
\State \textbf{until} convergence
\end{algorithmic}
\end{algorithm}

\newpage
\section{Experiments}\label{experiments}

\subsection{Datasets}

We start with experimentation on simple tasks rather than with more high-level tasks like natural language generation. 

Thus, we consider a dataset with simple grammar similar to \cite{kusner2016gans} with just one rule: alternating between the symbol "x" and operators +,-,*,/. Vocabulary size in this case is five, and we build two training sets of size 10,000 with sequences of length of respectively three and fifteen. The following shows some training examples in both lengths:\\

\begin{tcbraster}
\begin{tcolorbox}[height=2.4cm, width=3cm, title={\hspace{1cm} Sequence Length Three }]
\tab x-x \\
\tab x+x \linebreak
\tab x*x
\end{tcolorbox}
\begin{tcolorbox}[width=3cm, title={\hspace{1cm} Sequence Length Fifteen}]
\tab x/x*x/x*x-x*x+x \\
\tab x+x*x-x/x-x+x-x \linebreak
\tab x/x-x+x+x*x/x-x
\end{tcolorbox}
\end{tcbraster}

\subsection{Evaluation}

To evaluate the quality of generated sequences, we slide a window of length three over the sequence and count the proportion of correct sub-sequences which we refer to as Goodness Score. Such sub-sequences are of the form operator-x-operator or x-operator-x. A perfect Goodness Score is 1 and 13 for our datasets with sequence length three and fifteen respectively.

We also plot the Log Variance of different estimated gradients with respect to a network parameter of the generator as shown in \ref{seq15:var} and \ref{seq3:var}

\subsection{Results}
All the experimental results and graphs in the sections to follow use the following network architectures: 
\begin{enumerate}
\item Generator: many-to-many, unidirectional, one-layer LSTM
\item Discriminator: many-to-one, unidirectional, one-layer LSTM
\item Control Variate: Convolutional neural network.
\end{enumerate}
\begin{figure}[!htbp]
\begin{subfigure}[b]{0.49\textwidth}
    \includegraphics[width=\textwidth]{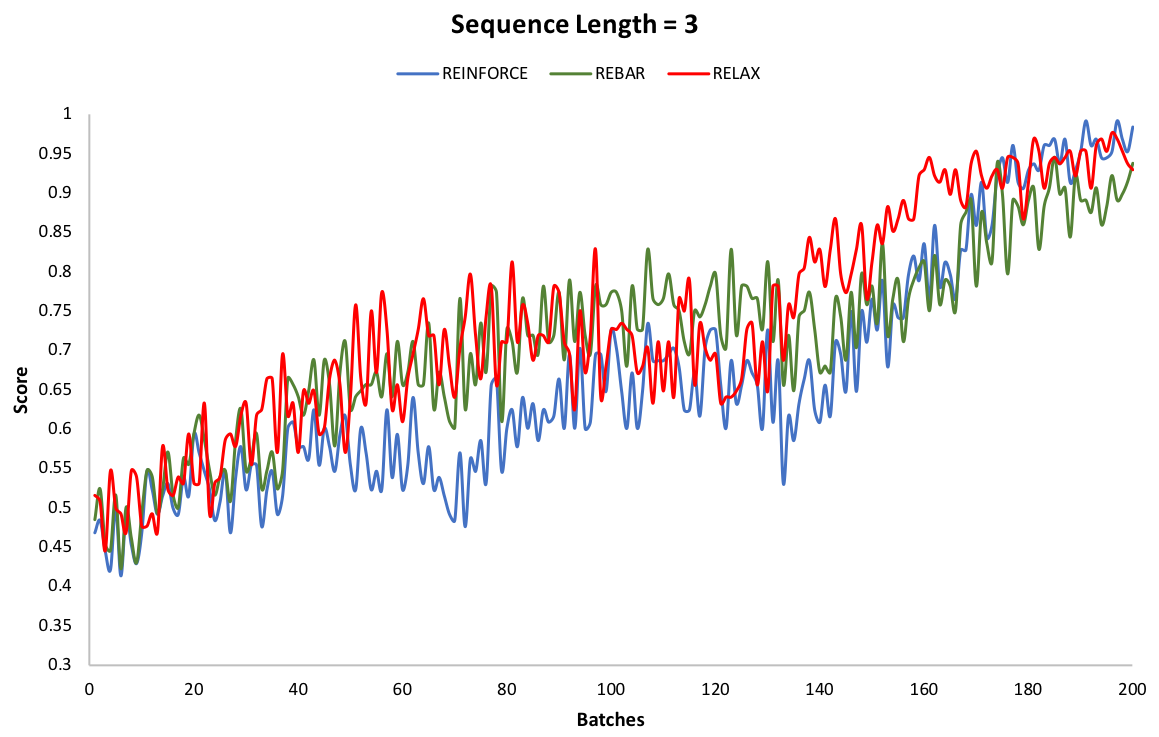}
    \caption{Goodness Score.}
    \label{seq3:score}
  \end{subfigure}
  \hfill
  \begin{subfigure}[b]{0.49\textwidth}
    \includegraphics[width=\textwidth]{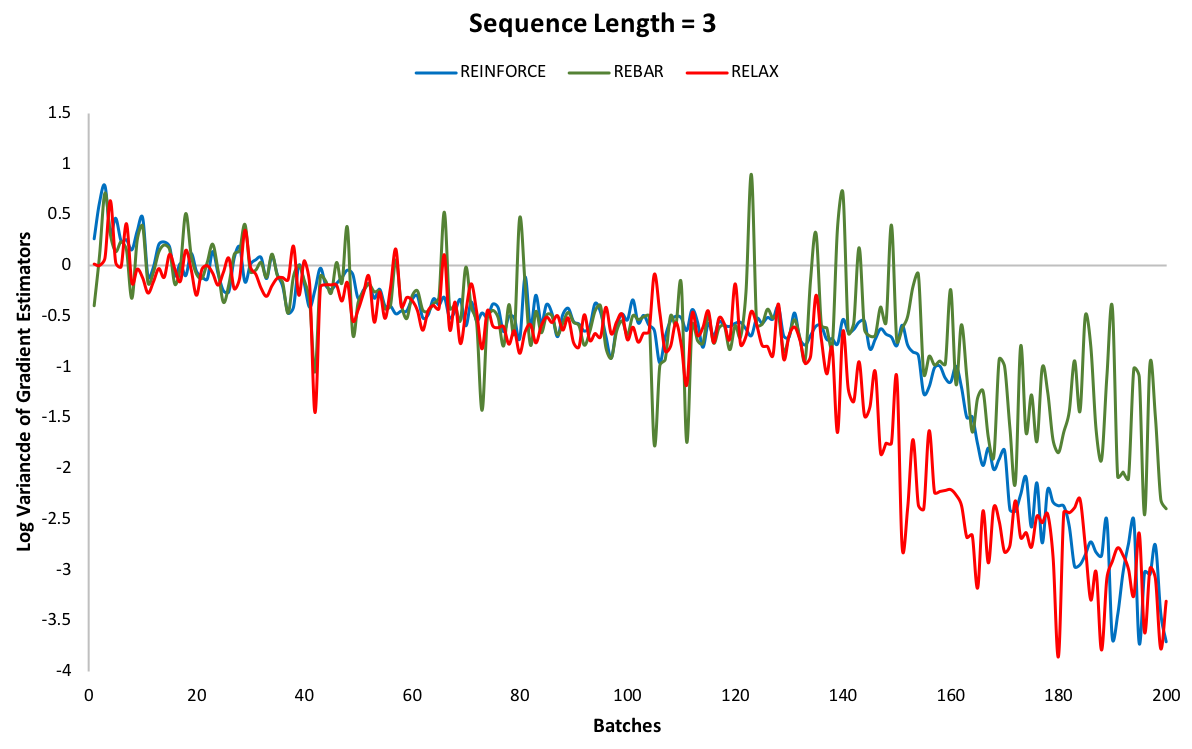}
    \caption{Log Variance}
    \label{seq3:var}
  \end{subfigure}
\caption{Experiment Results with Sequence Length Three}\label{seq-len-3}
\end{figure}

\begin{figure}[!htbp]
\begin{subfigure}[b]{0.49\textwidth}
    \includegraphics[width=\textwidth]{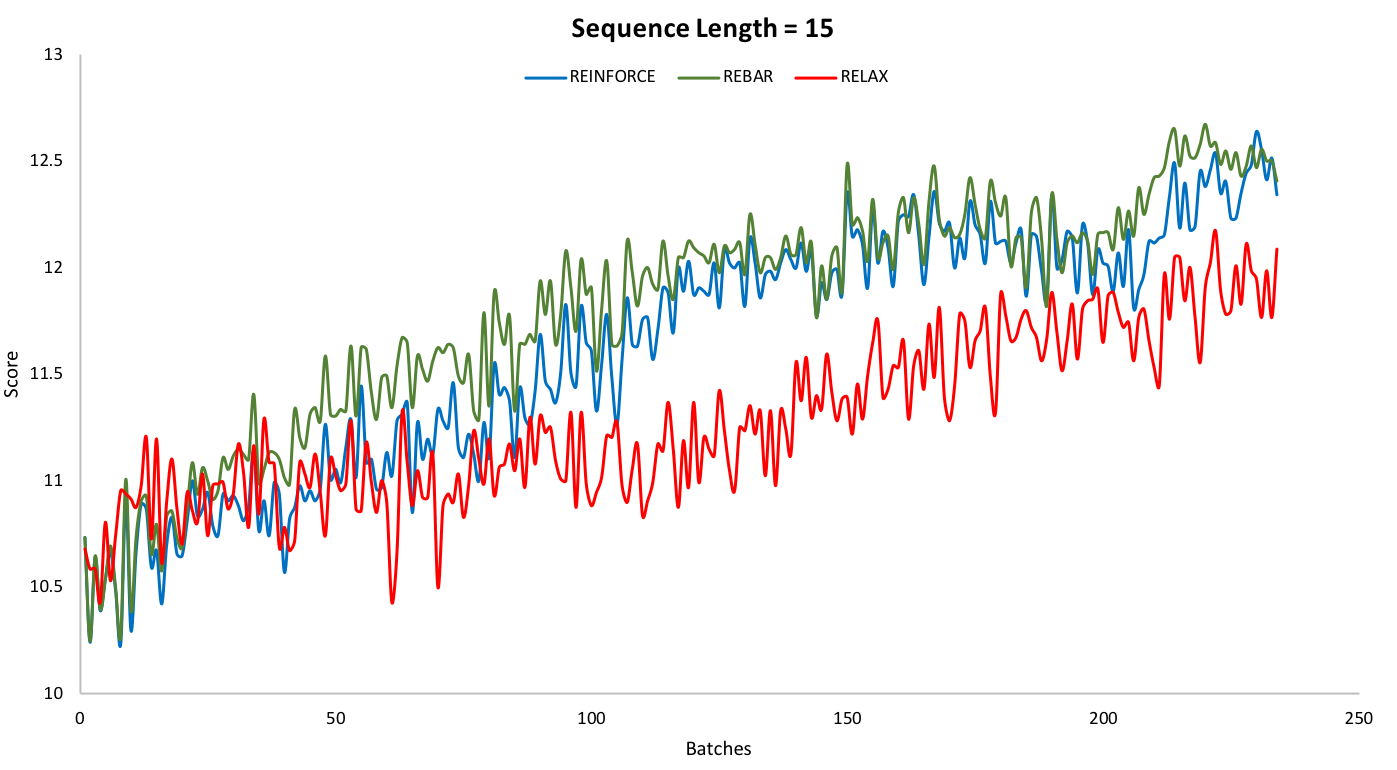}
    \caption{Goodness Score.}
    \label{seq15:score}
  \end{subfigure}
  \hfill
  \begin{subfigure}[b]{0.49\textwidth}
    \includegraphics[width=\textwidth]{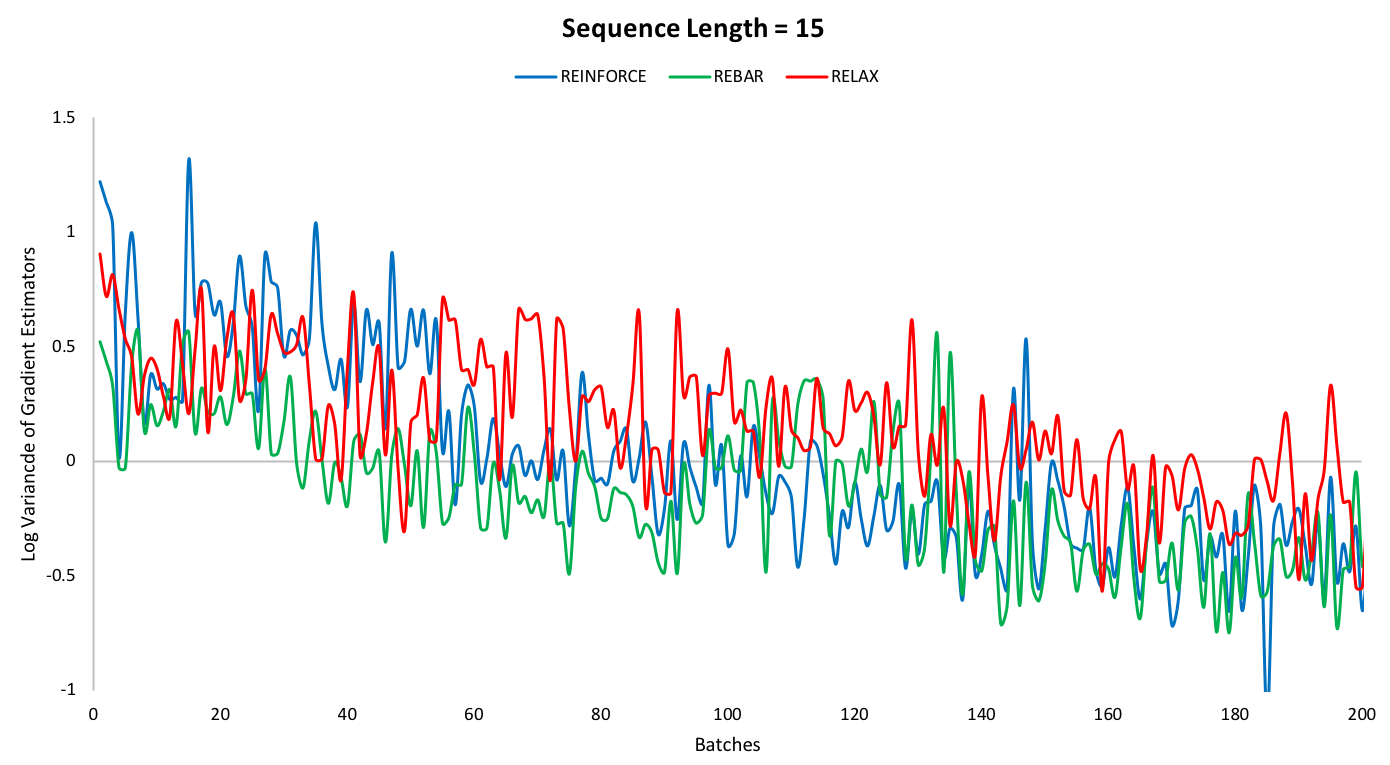}
    \caption{Log Variance}
    \label{seq15:var}
  \end{subfigure}
\caption{Experiment Results with Sequence Length Fifteen}\label{seq-len-15}
\end{figure}
We performed experiments to generate sequences of length three and fifteen using REINFORCE, REBAR and RELAX gradient estimators. 

%Table
The figures \ref{seq3:score} and \ref{seq15:score}, show a plot of the goodness score for the generated sequences. The figures \ref{seq3:var} and \ref{seq15:var} show the variance of the gradient of a parameter in the generator network over training epochs (in log units). We chose the parameter from the last layer of the generator. We use these plots to quantify the results of the GAN training.

We can observe in figure \ref{seq-len-3} that for generation of sequences of length three, RELAX produces better results and seems to converge faster. The Log Variance plot also shows lower variance in RELAX compared to that of REBAR and REINFORCE.

Whereas for sequences of length fifteen, we observed results of RELAX to be slightly inferior to REINFORCE. We also found the variance to be very noisy. These results are inconclusive.

\subsubsection{Generated Data}
The frequency counts of individual tokens (operators and x) are almost the same as that in the original dataset for all three gradient estimators. For example, in sequences with length three, frequency count-based unigram probability of 'x' is close to 0.6, while that of  '+' and '-' are close to 0.085 in a batch of generated sequences . These are quite similar to the original unigram probabilities of 0.67 for 'x' and 0.083 for each operator in the original toy dataset. Based on this and on direct observations of the generated sequences, we conclude that there is no mode collapse. We show below sequences generated using RELAX gradient estimator in comparison to text generated using baseline gradient estimator, REINFORCE.

\begin{tcbraster}
\begin{tcolorbox}[width=3cm, title={\hspace{1cm} RELAX Generated Text}]
\tab +x-x-x+x+x-x/x+ \\
\tab /x+x-x/x+x-x+x/ \linebreak
\tab /xxx/x+x/x*x-x- \\
\tab *x-x*x/x+x/x+x+
\end{tcolorbox}
\begin{tcolorbox}[width=3cm, title={\hspace{0.7cm} REINFORCE Generated Text}]
\tab x/x-xx/x+x+xx+x \linebreak
\tab -x*x*x/x+x-x/x+ \linebreak
\tab *xxx/x*x-x*x/x+ \\
\tab +x+x/x-x*x-x-xx
\end{tcolorbox}
\end{tcbraster}

\section{Limitations and Future Work}\label{limitations}
% \subsection{Limitations}
$\bullet$ \textbf{Pre-training}: Similarly to SeqGAN \cite{yu2017seqgan}, we also find pre-training to be of significant importance in the success of the subsequent adversarial training. While SeqGAN needs 120 epochs of pre-training for their experiments on the Obama speech dataset, our simpler learning tasks only require a few epochs of pre-training. 

Interestingly, the more complex the task, the longer pre-training seems to be necessary. In sequences of length three, we pre-train for one epoch, and in sequences of length fifteen, for three epochs. Our experiments show that a minimal amount of pre-training is needed to kickstart adversarial training. And without any pre-training, our gradient estimators simply cannot learn. 

$\bullet$ \textbf{Scalability}: As we have seen, it is possible to successfully train a GAN to generate discrete sequences, providing initial pre-training of the generator. However, our experiments were done on datasets characterized by a simple pattern to learn. Backpropagation through time and gradient variance computation are both computational bottlenecks that slow down training as sequence length and training set length increase. The next step of our study would be to train our model on textual datasets, and use natural language processing metrics like BLEU. 

$\bullet$ \textbf{Variance display}: RELAX involves training a neural network to minimize the gradient variance. Doing so, we backpropagate the L2-norm of this gradient through the annex neural network. But displaying the variance of this gradient for all the parameters in the generator would be tedious. In our study, we picked the first parameter of the last layer. A more rigorous alternative would involve checking this variance for every parameter ; or averaging. 

\section{Conclusion}\label{conclusion}
In this work, we have pioneered the application of recently developed unbiased and low-variance gradient estimation algorithms REBAR and RELAX to sequence generation using generative adversarial training. Our study systematically compared the REINFORCE, REBAR, and RELAX estimators, in the hope of establishing benchmarks for these methods. We field tested the convergence of these estimators on two synthetic sequence datasets with a simple grammar and evaluation metric. We did not observe any mode collapse, and saw promising results with a sequence length of three. \\

However, our results were not sufficient to fully assert the superiority of one specific gradient estimation algorithm over the REINFORCE baseline. We would like to conduct in-depth tests and broad sets of experiments as an extension to this work. Our future work would also involve applying  these gradient estimators to more high-level tasks like natural language generation (NLG). 

\newpage
\bibliographystyle{plain}
%\bibliography{bib}

%******************** APPENDIX START ************************%
\newpage
\appendixpage
\appendix
\section{Background: Gradient Estimators} \label{appendix:background-gradient-estimators}
In this section, we provide a quick refresher of the different gradient estimators for the expectation of function with discrete variables. The problem in focus is of estimating the gradient of a function of the form:

\begin{equation}
J(\theta) = \mathbb{E}_{p(b|\theta)}[f(b)]
\end{equation}
For more convenience, we use the following convenient abbreviations for gradient estimators: RF for REINFORCE, RB for REBAR, RX for RELAX
\subsection{REINFORCE}

First introduced by Ronald J. Williams in 1992 \cite{williams1992simple}, the REINFORCE gradient estimator is expressed as:
\begin{equation}
\nabla_{RF}J(\theta) = f(b)\frac{\partial}{\partial \theta}\log(p(b|\theta))
\end{equation}
Although simple to compute and unbiased, it suffers from high variance \cite{greensmith2004variance}. In the context of a very sensitive GAN training \cite{lucic2017gans}, we believe it to be a major drawback. 

\subsection{REBAR}

Introduced in 2017 by Tucker et al. \cite{tucker2017rebar}, the REBAR gradient estimator leverages two simple calculus tricks. The first one is the re-parameterization trick \cite{kingma2015variational}, where the discrete latent variable can be written as a differentiable, deterministic function of a sample from a fixed distribution:
\begin{equation}
\nabla_{reparam}J(f) = \frac{\partial f}{\partial b} \frac{\partial b}{\partial \theta}
\end{equation}
The second trick is to make use of control variates. In order to reduce variance of the gradient estimates, we subtract from the gradient estimator a simple function of the variable \(b\). We ensure unbiasedness by adding back the expected value of this function. This trick is written as:
\begin{equation}
\nabla_{new}J(\theta) = \nabla_{old}J(\theta) - c(b) + \mathbb{E}_{p(b)}[c(b)] 
\end{equation}
Leveraging these two ideas, the REBAR estimator full equation is:
\begin{equation}
\nabla_{RB}J(\theta) = \mathbb{E}_{p(u,v)}[[f(H(z)) - \eta f(\sigma_{\lambda}(\tilde{z}))]\frac{\partial}{\partial \theta}\log(p(b))|_{b=H(z)} + \eta \frac{\partial}{\partial \theta}f(\sigma_{\lambda}(z)) - \eta \frac{\partial}{\partial \theta}f(\sigma_{\lambda}(\tilde{z}))]
\end{equation}
where $z$ and $\tilde{z}$ are continuous relaxations of the discrete variable b as shown in \cite{grathwohl2017backpropagation}.\\

In practice, we take a single-sample Monte-Carlo estimation of this expected value. REBAR introduces two new hyper-parameters $\eta$ and $\lambda$. For our experiments, we set $\eta$ to 1, and tune the temperature parameter $\lambda$. REBAR is unbiased and lowers variance compared to REINFORCE.

\subsection{RELAX}

Introduced in 2017 by Grathwohl et al. \cite{grathwohl2017backpropagation}, RELAX generalizes REBAR by letting the control variate be a free form function, like a neural network for instance. The equation for RELAX is:
\begin{equation}
\nabla_{RX}J(\theta) = [f(b) - c_{\phi}(\tilde{z})]\frac{\partial}{\partial \theta}\log(p(b|\theta)) + \frac{\partial}{\partial \theta}c_{\phi}(z) - \frac{\partial}{\partial \theta}c_{\phi}(\tilde{z})
\end{equation}
where $c_{\phi}$ is a general control variate. \\
It can for instance be a neural network trained to minimize the variance of the gradient. \cite{grathwohl2017backpropagation} exhibits the following convenient expression of the gradient of this variance:
\begin{equation}
\frac{\partial}{\partial \phi}Variance(\nabla_{estimator}J(\theta)) = \mathbb{E}[\frac{\partial}{\partial \phi}(\nabla_{estimator}J(\theta))^{2}]
\end{equation}

\section{Temperature tuning}

The REBAR and RELAX estimators both require to set a temperature parameter $\lambda$. This parameter can have a significant impact on quality of training and stability. The following shows experimental results for different order of magnitudes of $\lambda$ in REBAR applied to sequences of length three:

\begin{figure}[!htbp]
\includegraphics[width = 0.5\textwidth]{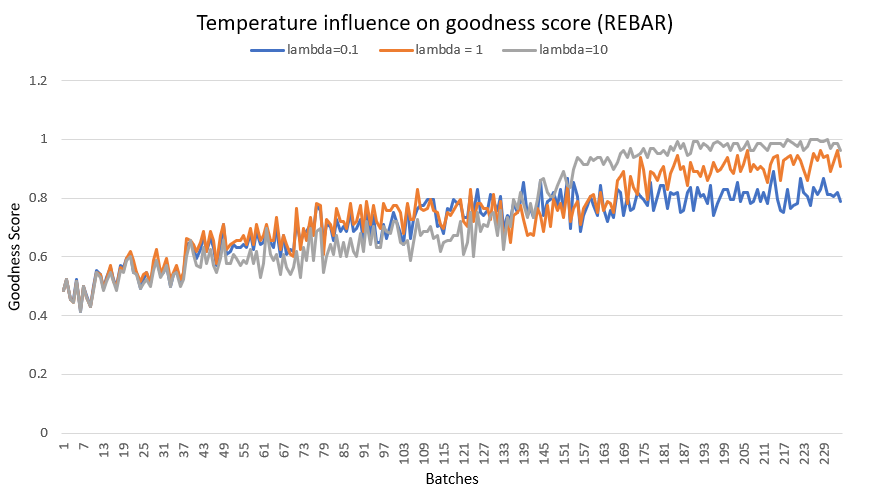}
\includegraphics[width = 0.5\textwidth]{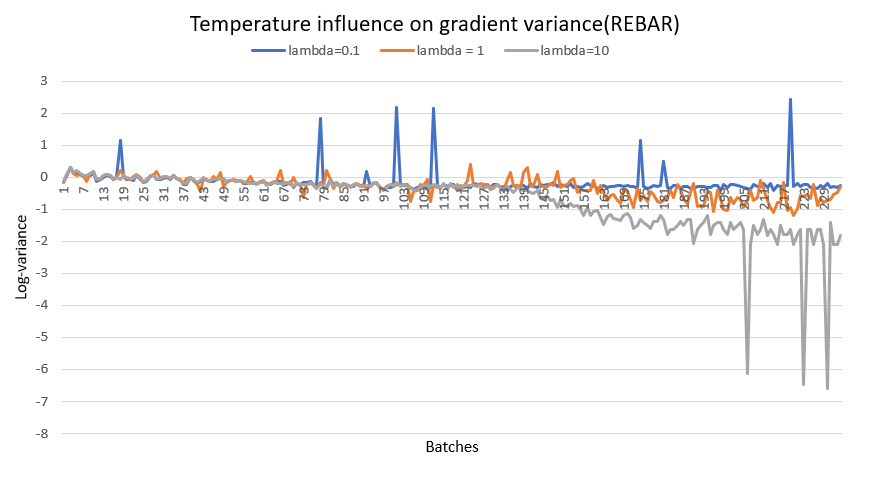}
\caption{REBAR trained on sequences of length three with different $\lambda$ values}
\end{figure}

In all our other experiments, we used $\lambda=1$. 

% \section{Methods and Materials}
% \label{appendix:methods-and-materials}
% \subsection{Appendix subsection.}

%******************** APPENDIX END ************************%

\end{document}